\documentclass{article}

\usepackage{PRIMEarxiv}

\usepackage[utf8]{inputenc} 
\usepackage[T1]{fontenc}    
\usepackage{hyperref}       
\usepackage{url}            
\usepackage{booktabs}       
\usepackage{amsfonts}       
\usepackage{nicefrac}       
\usepackage{microtype}      
\usepackage{lipsum}
\usepackage{fancyhdr}       
\usepackage{graphicx}       
\graphicspath{{media/}}     
\usepackage{amsmath} 
\usepackage{multirow}

\title{\textbf{CPSD}Bench: A Large Language Model Evaluation Benchmark and Baseline for \textbf{C}hinese \textbf{P}ublic \textbf{S}ecurity \textbf{D}omain
\thanks{\textit{\underline{Citation}}: 
\textbf{Tong, Xin, et al. "CPSDBench: A Large Language Model Evaluation Benchmark and Baseline for Chinese Public Security Domain." arXiv preprint arXiv:2402.07234 (2024).}} 
}

\author{
  Xin Tong \\
  School of Information and Network Security \\
  People's Public Security University of China \\
  Beijing\\
  \texttt{tongxindotnet@outlook.com} \\
   \And
  Bo Jin* \\
 National Engineering Research Center of Classified Protection and Safeguard Technology for Cybersecurity\\
  The Third Research Institute of the Ministry of Public Security of China \\
  Shanghai\\
  \texttt{jinbo@gass.cn} \\
   \And
  Zhi Lin \\
  Department of Engineering Physics\\
   Tsinghua University\\
   \&\\
  Science and Technology Information Bureau \\
  The Ministry of Public Security of China \\
  Beijing\\
  \texttt{LZ400LZ400@163.com} \\
  \And
  Binjun Wang \\
  School of Information and Network Security \\
  People's Public Security University of China \\
  Beijing\\
  \texttt{wangbinjun@ppsuc.edu.cn} \\
   \And
  Ting Yu, Qiang Cheng \\
  The Third Research Institute of the Ministry of Public Security of China
  Shanghai\\
  \texttt{rainydaily0@163.com herculescheng@qq.com} \\
}

\begin{document}
\maketitle

\begin{abstract}
Large Language Models (LLMs) have demonstrated significant potential and effectiveness across multiple application domains. To assess the performance of mainstream LLMs in public security tasks, this study aims to construct a specialized evaluation benchmark tailored to the Chinese public security domain—CPSDbench. CPSDbench integrates datasets related to public security collected from real-world scenarios, supporting a comprehensive assessment of LLMs across four key dimensions: text classification, information extraction, question answering, and text generation. Furthermore, this study introduces a set of innovative evaluation metrics designed to more precisely quantify the efficacy of LLMs in executing tasks related to public security. Through the in-depth analysis and evaluation conducted in this research, we not only enhance our understanding of the performance strengths and limitations of existing models in addressing public security issues but also provide references for the future development of more accurate and customized LLM models targeted at applications in this field.
\end{abstract}


\section{Introduction}
In the domain of artificial intelligence, significant advancements have been made in text comprehension and generation by large language models such as the ChatGPT series, LLaMA\cite{touvron2023llama,touvron2023llama2}, and other models based on deep learning. The efficiency and multifunctionality of these models have led to their widespread adoption across various industries, particularly in fields requiring advanced language understanding and processing capabilities. public security work, a task with high demands for information processing and analysis, is increasingly reliant on these advanced technologies. The potential application of large language models is especially notable in areas such as criminal investigation, detection of telecommunication fraud, and analysis of social sentiment.

Despite large language models being proven to excel in various domains such as law, finance, and medicine, it remains an open question whether these models can achieve desired outcomes in specific areas, such as public security work. The uniqueness of public security work lies in its high demands for accuracy and reliability, and the often more sensitive and complex nature of the data handled. Furthermore, most existing language models are primarily designed for general language processing tasks, rather than for specific applications in the field of public security. Therefore, evaluating and optimizing these models to cater to the unique requirements of the public security domain becomes particularly crucial.

The objective of this research is to establish a standard framework for evaluating and comparing the performance of mainstream large language models in tasks related to public security work. Through this framework, we aim to explore the effectiveness of large language models in specific tasks such as telecommunication fraud detection, sentiment analysis, and rumor detection. This study not only aids in understanding and evaluating the strengths and limitations of existing models in handling public security-related tasks but also provides insights and guidance for the future development of customized models more suited for public security applications.
\begin{itemize}
\item \textbf{Constructed a Dataset for Public Security Tasks.} We have developed a dataset tailored for public security tasks. CPSDBench, a specialized evaluation benchmark dataset designed explicitly for public security tasks, was constructed. By gathering data related to public security from real-world scenarios, we have assembled a comprehensive set of test datasets from four dimensions: text classification, information extraction, question answering, and text generation. This dataset not only encompasses key tasks in the public security domain but also simulates complex situations encountered in the real world.

\item \textbf{Selection and Design of Appropriate Evaluation Metrics.} Considering the uniqueness of different tasks, we meticulously selected and designed a series of evaluation metrics. Especially for information extraction tasks, we introduced an innovative two-stage evaluation metric aimed at more accurately quantifying the performance of LLMs in identifying and extracting key information.

\item \textbf{Evaluation of Mainstream Open-source and Proprietary LLMs.} Our research, through a comprehensive evaluation of nine mainstream LLMs currently available on the market, provides valuable insights into the potential application of these models in the public security domain. The evaluation included proprietary models such as GPT-4 and ChatGLM-4, as well as open-source models like Atom and XVerse. Through this extensive assessment, we not only showcased the performance of each model in public security tasks but also revealed their respective strengths and limitations, offering guidance for future research and development.
\end{itemize}

\section{Related Work}
\subsection{Large Language Model}
Large language models represent a significant innovation in the field of Natural Language Processing (NLP) in recent years. These models are typically composed of billions or even trillions of parameters, enabling them to understand and generate human language. Compared to traditional NLP models, large language models possess a deeper capacity for language comprehension and generation, capable of handling more complex linguistic tasks.

In terms of model training, the training process of LLMs is more complex compared to other deep learning models. Specifically, the training of LLMs usually encompasses three stages: pre-training, supervised fine-tuning, and reinforcement learning based on human feedback.

During the pre-training phase, the model is trained on large-scale text datasets to learn the fundamental rules and structures of language. These text datasets typically include a variety of texts collected from the internet, such as books, articles, and web pages, covering a wide range of topics and styles. The aim of pre-training is to equip the model with a broad knowledge and understanding of language. In this phase, the model learns to predict the next word in a text (i.e., the language modeling task)\cite{radford2018improving}.

Following pre-training, large language models enter a second critical phase – supervised fine-tuning, also known as instruction tuning. The aim of this stage is to enhance the model's ability to understand and respond to specific user commands or queries. Typically, the model is trained with a series of samples containing explicit instructions. Each sample includes a command (for instance, answering a question, writing a text, performing a translation) and the corresponding ideal output. This training approach augments the model's responsiveness to specific task instructions, making it more precise and effective in practical applications.

To ensure that LLMs align with human preferences and values, Reinforcement Learning from Human Feedback (RLHF) strategies are often incorporated during the training process. RLHF employs reinforcement learning algorithms, such as Proximal Policy Optimization\cite{schulman2017proximal}, to adapt LLMs to human feedback. This method integrates human judgment into the training loop, aiming to develop LLMs that are better aligned with human values, exemplified by models like InstructGPT\cite{ouyang2022training}.

One of the core characteristics of LLMs (Large Language Models) is their emergent ability\cite{wei2022emergent}, which enables the models to exhibit human-like cognition and understanding capabilities in handling complex linguistic tasks. Specifically, this emergent ability is manifested in three key areas: contextual learning, instruction following, and chain-of-thought.

\begin{itemize}
\item[1)]
Contextual Learning: Large language models demonstrate exceptional proficiency in contextual learning\cite{brown2020language}. This means the models can understand and utilize the contextual information of given text to generate relevant and coherent responses or texts. For instance, in a conversational system, the model responds not just to individual inputs but also makes more rational and coherent replies based on the content of previous dialogues. This capability stems from the model's analysis of vast quantities of textual data during training, enabling it to capture the subtle connections and deeper meanings within language.

\item[2)]
Instruction Following: Instruction following is another significant characteristic of large language models. These models can comprehend user instructions and generate outputs that conform to these instructions. This ability renders the models highly flexible in application, adaptable to a variety of tasks ranging from simple query responses to complex creative writing. For example, when a user requests the model to generate an article on a specific topic or answer a complex question, the model can accurately understand these instructions and produce corresponding high-quality content.

\item[3)]
Chain-of-Thought: Chain-of-Thought (CoT)\cite{wei2022chain} refers to the model's ability to demonstrate logic and continuity in text generation, akin to the human thinking process. This is manifested not only in answering individual questions but also in the model's capacity to maintain consistency and logical coherence across a series of interactions. This capability enables the model to perform as though it possesses genuine understanding and cognitive abilities when handling complex dialogues or executing tasks that require multi-step reasoning.
\end{itemize}

\subsection{Application and Evaluation Benchmarks of LLMs}
Currently, the field of LLMs is undergoing rapid development. On one hand, a variety of general-purpose large models, such as ChatGPT, Claude, ChatGLM\cite{du2022glm, zeng2022glm}, and Llama\cite{touvron2023llama,touvron2023llama2}, have emerged, demonstrating their broad applicability across multiple tasks and domains. On the other hand, there is a swift growth in large models tailored for specific application domains. This includes BloombergGPT\cite{wu2023bloomberggpt} applied in the financial sector, ChatLaw\cite{cui2023chatlaw} focused on the legal domain, EduChat\cite{dan2023educhat} oriented towards educational tasks, and BioMedGPT\cite{zhang2023biomedgpt} and HuaTuo\cite{wang2023huatuo, li2023huatuo} for the biopharmaceutical field. These models not only exemplify the depth of customization and optimization of large language models in specific vertical domains but also highlight their practical application potential and value within these areas.

With the advancement of LLMs technology, the need to evaluate their performance and capabilities has increasingly become critical, impacting not only the technical sphere but also commercial decision-making and public perception. The evaluation of LLMs typically involves multiple dimensions, including the model's general capability, generalizability, robustness, cross-domain performance, multilingual ability, interpretability, and safety. Mainstream benchmarks, such as SuperGLUE\cite{wang2019superglue}, C-Eval\cite{huang2023c}, and MMLU\cite{hendrycks2020measuring}, often focus on assessing the comprehensive effectiveness of LLMs. However, these benchmarks typically utilize general task data, making it challenging to evaluate the performance of LLMs in specific domains (like medicine, law).

In this context, a series of domain-specific evaluation benchmarks have been proposed. They challenge LLMs to demonstrate not only general language comprehension but also the ability to accurately process and generate information specific to particular domains.

In the medical domain, MultiMedQA\cite{singhal2023large} is a medical QA benchmark focusing on medical examinations, medical research, and consumer health issues. It encompasses seven datasets related to medical quality assurance. The aim of this benchmark is to assess the performance of LLMs in clinical knowledge and QA capabilities. CMB\cite{wang2023cmb} is a comprehensive, multi-level Chinese medical benchmark. It contains 280,839 questions and 74 complex case consultation questions, covering all clinical medical specialties and various professional levels. This platform is designed to thoroughly evaluate large models' medical knowledge and clinical consultation abilities. PromptCBlue\cite{zhu2023promptcblue} is a benchmark for evaluating LLMs in Chinese medical scenarios, employing a vast corpus of Chinese medical texts for testing. These texts include medical records, medical literature, and pharmaceutical instructions, covering a broad range of medical fields. PromptCBlue can be used to assess the performance of LLMs in identifying key information such as disease names, medication names, and symptom descriptions.

In the legal domain, Fei\cite{fei2023lawbench} introduced Lawbench to precisely evaluate the legal capabilities of LLMs. This benchmark assesses on three cognitive levels: legal knowledge recall, legal knowledge comprehension, and legal knowledge application. Dai\cite{dai2023laiw} proposed the LAiW benchmark for the Chinese legal domain, encompassing dimensions of evaluation for LLMs that include basic legal NLP capabilities, basic legal application abilities, and complex legal application abilities. To further assess LLMs in multilingual legal tasks, Niklaus\cite{niklaus2023lextreme} constructed LEXTREME, a benchmark covering 11 legal-related datasets in 24 languages.

In the financial domain, FinanceBench\cite{islam2023financebench} is the first test suite designed to evaluate the performance of LLMs in open-book financial Question Answering (QA). It comprises 10,231 questions about publicly listed companies, along with corresponding answers and evidence strings, and includes tests results for baselines such as GPT-4, Llama2, and Claude2. FinEval\cite{zhang2023fineval} is a benchmark specifically designed for LLMs in the Chinese financial domain. It includes 4,661 questions covering content related to Finance, Economy, Accounting, and Certificates. Lei\cite{lei2023cfbenchmark} proposed CFBenchmark to evaluate the performance of Chinese financial assistant LLMs. The basic version of CFBenchmark aims to evaluate fundamental capabilities in processing Chinese financial texts from three aspects: recognition, classification, and generation.

\section{Method}
In China, public security constitutes the core of police work, encompassing critical areas such as crime prevention, case investigation, intelligence analysis, and the maintenance of public order. This significant responsibility is typically borne by various specialized police forces to ensure targeted and efficient maximization. For instance, narcs play an indispensable role in drug control efforts, with their primary tasks being the identification of individuals, locations, and items related to drug activities, thereby advancing drug control initiatives. Cyber police focus on maintaining order in cyberspace, analyzing and assessing the public's emotional responses to social events, timely identifying and dealing with online rumors, and ensuring the stability and safety of the online environment. criminal police, security officers, and traffic police play crucial roles in the investigation of criminal cases, handling civil disputes, and the control of traffic violations, respectively. Furthermore, in response to the severe challenges posed by the emerging crime of telecommunications fraud, the criminal police department has established an anti-fraud center. Its main tasks include quickly filtering potential fraudulent information, effectively issuing warnings, and blocking telecommunications fraud activities to protect the public's financial safety and reduce economic losses. This collaborative division of labor ensures the comprehensiveness and professionalism of public security work, providing a solid foundation for maintaining social stability and citizen safety.

Despite the continuous emergence of domain-specific evaluation benchmarks, a significant gap persists in assessing the application of LLMs in the public security domain. Tasks in public security, such as sentiment analysis, police incident classification, and fraud detection, are not only diverse in nature but also highly complex and unique in terms of difficulty and data processing methodologies. Existing general language model evaluation benchmarks struggle to comprehensively and accurately measure the capabilities of LLMs in handling these specific public security tasks. Furthermore, the accuracy and reliability requirements for LLMs in the public security domain are exceedingly stringent. For instance, minor errors in fraud detection and police incident classification tasks can lead to significant consequences. Current evaluation benchmarks may not adequately consider these high-standard demands. Additionally, data in public security tasks may include sensitive and complex factors like adversarial perturbations\cite{zhang2019adversarial}, violent or terroristic content, and discriminatory materials, all of which could significantly impact the performance of LLMs.

Therefore, the development of a specialized benchmark for the public security domain is crucial. This not only more effectively measures model performance in complex scenarios but also ensures that model performance meets rigorous requirements in high-risk environments. This is vital for aiding the development and optimization of LLMs more suitable for public security applications and can significantly enhance the capabilities of public security departments in handling complex situations. Concurrently, this also aids in advancing the application and development of technology in the public security field. To this end, we have developed CPSDBench — a benchmark specifically designed for evaluating LLMs in the Chinese public security domain, aiming to provide a comprehensive and precise assessment of model performance in this area.

\subsection{Tasks and Datasets}
\begin{figure}
  \centering
  \includegraphics[scale=0.39]{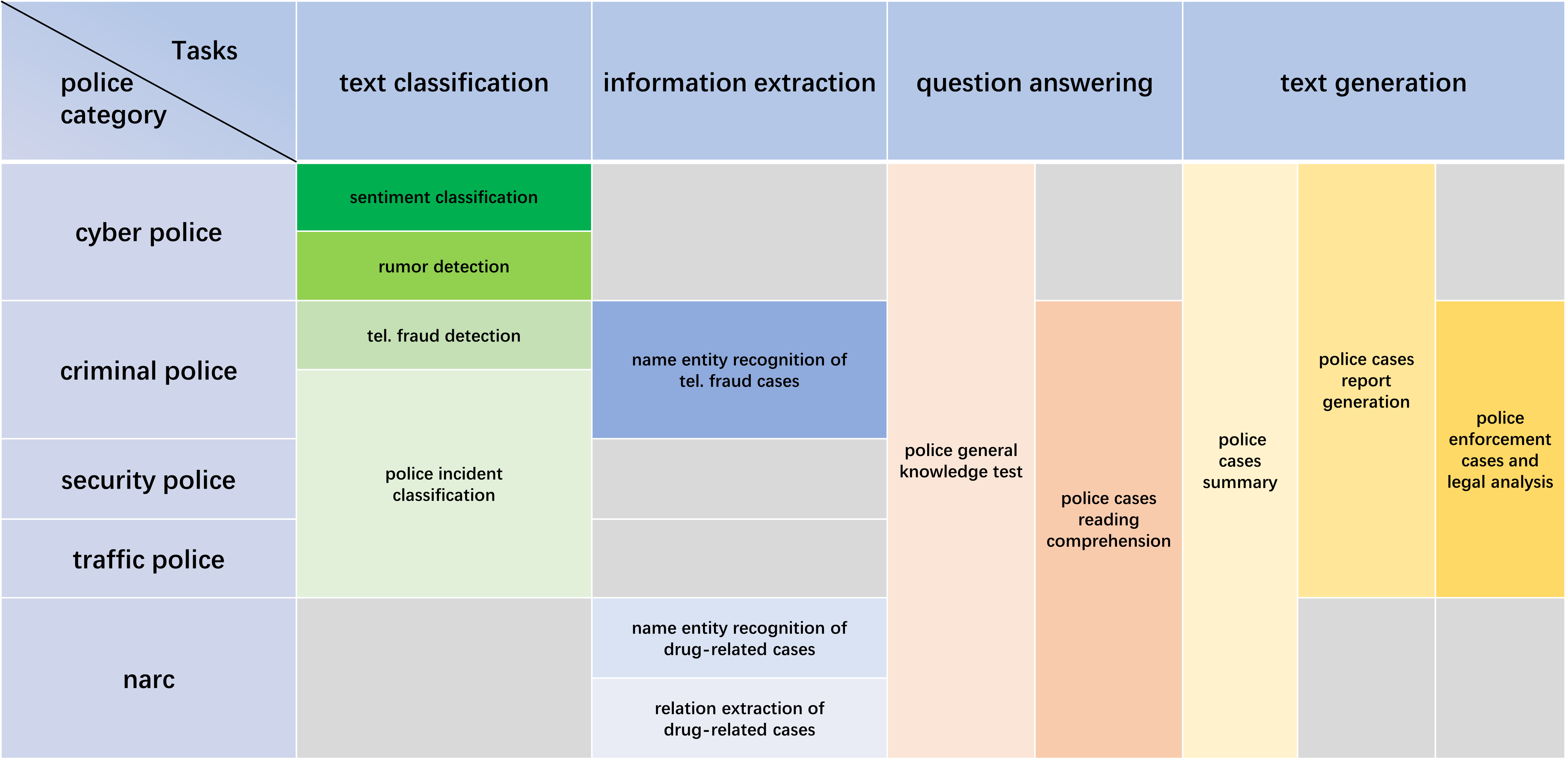}
  \caption{The evaluative content of CPSDBench and its correspondence to the police category in China. For various police categories, the CPSDBench benchmark testing framework has designed at least five types of tasks, aiming to achieve a comprehensive and specific assessment of each category, ensuring the evaluation is both holistic and targeted.}
  \label{fig:fig1}
\end{figure}

To evaluate the application performance of LLMs in the public security domain thoroughly and comprehensively, CPSDBench has constructed a four-dimensional assessment task system, encompassing key dimensions such as text classification, information extraction, question-answering, and text generation. This assessment system closely revolves around actual public security business scenarios, aggregating a diverse dataset (as shown in Table~\ref{tab:table1}), aiming to conduct a more detailed and comprehensive exploration of LLMs' abilities in parsing and generating public security-related data. Furthermore, the correlation between the evaluation content adopted by CPSDBench and the business domains of the Chinese police is graphically presented (Figure~\ref{fig:fig1}). Considering the high costs associated with utilizing commercial-grade large language models (e.g., ChatGPT, ChatGLM-130B, etc.), the scale of each dataset is generally controlled between tens to hundreds of samples to ensure the evaluation's economic viability and operability.

\begin{table}
 \caption{The evaluation dimensions and task details of CPSDBench. This benchmark framework primarily conducts an in-depth assessment of the capabilities of LLMs in the Chinese public security domain, covering four core dimensions: text classification, information extraction, question-answering, and text generation.}
  \centering
  \begin{tabular}{llll}
    \toprule
    Evaluation Dimensions  & Tasks  & Samples & Category Information \\
    \midrule
    \multirow{4}*{Text Classification} 
            & Sentiment Classification & 600 & 3 classes \\
            & Rumor Detection & 400 & 2 classes \\
            & Telecommunication Fraud Detection & 400 & 2 classes \\
            & Police Incident Classification & 400 & 4 classes \\
            \cmidrule(r){2-4}
    \multirow{3}*{Information Extraction} 
            & Drug-Related Case NER & 200 & 5 types of entities \\
            & Drug-Related Case RE & 200 & 4 types of relationships \\
            & Telecommunication Fraud Case NER & 200 & 6 types of entities \\
            \cmidrule(r){2-4}
    \multirow{2}*{Question Answering} 
            & Police Cases Reading Comprehension & 200  & 3 types of questions \\
            & Police General Knowledge Test & 200  & 3 types of questions \\
            \cmidrule(r){2-4}
    \multirow{2}*{Text Generation} 
            & Police Cases Summary & 200 & N/A \\
            & Police Cases Reports Generation & 100 & N/A \\
            & Police Enforcement Cases and Legal Analysis  & 50 & N/A \\
    \bottomrule
  \end{tabular}
  \label{tab:table1}
\end{table}

\subsubsection{Text Classification}
Text classification, a key component of NLP, aims to categorize text data into predefined categories or labels. This process is crucial for organizing, summarizing, and retrieving vast amounts of public security big data. To assess the capability of LLMs in handling Chinese public security domain text classification tasks, CPSDBench provides four text classification datasets directly sourced from real public security business scenarios. These datasets not only possess practical application value but also offer a tangible and specific standard for evaluating the performance of LLMs in this domain.

\begin{itemize}
\item[1)]
Sentiment Classification: Sentiment analysis, a foundational task in social sentiment analysis, is crucial for revealing and analyzing hot topics and related public opinion dynamics. By classifying the emotions of social media users, it is possible to effectively discern public emotional trends. CPSDBench has introduced a Weibo sentiment analysis dataset specifically for sentiment classification. This dataset is subdivided into three emotional categories: positive, negative, and neutral sentiments, each containing 200 samples. The aim is to comprehensively evaluate the capabilities of LLMs in the domain of sentiment recognition.

\item[2)]
Rumor Detection: Online rumors have become a significant challenge to cyberspace security and public security. Police departments often need to invest substantial resources in manual rumor identification and verification. Existing research demonstrates that deep learning models can effectively analyze the characteristic styles of rumors and identify them accordingly. CPSDBench provides a dataset specially designed to assess the capabilities of LLMs in rumor detection. This dataset includes two categories: rumors and non-rumors, with 200 samples in each category, to test the accuracy and robustness of models in this domain.

\item[3)]
Telecommunication Fraud Detection: As one of the core operations of cyber police departments, telecommunication fraud detection often faces data that is complex and adversarial in nature. CPSDBench includes a dataset for telecommunication fraud message detection, aimed at assessing the model's ability to identify fraudulent messages. This dataset comprises 200 fraudulent message samples and 200 regular message samples, covering various types of fraud such as banking and gambling scams.

\item[4)] 
Police Incident Classification: Police departments routinely handle a large volume of emergency reports, and efficient classification of these reports is crucial. CPSDBench has introduced a set of real police incident data from a police command center, with theobjective of evaluating the performance of LLMs in police incident classification. The dataset involves four types of incidents: theft, telecommunication fraud, traffic cases, and intentional injury cases. Furthermore, The dataset comprises a total of 400 samples, and the distribution of the data is balanced.

\end{itemize}

\subsubsection{Information Extraction}
The objective of information extraction is to automate the extraction of structured information from text, such as entities, relationships, and events. In the field of public security, this process is critical for rapidly and accurately distilling key data from voluminous case reports and incident information. To assess the performance of LLMs in this domain, CPSDBench provides two datasets specifically designed for information extraction.

\begin{itemize}
\item[1)]
Drug-Related Case Reports Information Extraction: This dataset contains report texts related to drug cases, intended for use in Named Entity Recognition (NER) and Relation Extraction tasks. For the NER task, the dataset includes five types of entities: names, dates, locations, drug types, and quantities of drugs. Notably, the data labels contain a large number of duplicate entities, making the evaluation process more stringent and complex. A prediction is considered entirely correct only when the text and quantity of an entity are accurately extracted. For the Relation Extraction task, the dataset encompasses four types of relationships: traffic\_in, sell\_drugs\_to, possess, and provide\_shelter\_for.

\item[2)]
Telecommunication Fraud Case Information Extraction: This dataset is specifically designed for the identification of named entities in telecommunications fraud cases. The current version encompasses six types of entities: the reporter, the time of the incident, the location of the incident, the amount involved, social media accounts, and telephone numbers. The design of this dataset aims to evaluate the accuracy and efficiency of LLMs in processing complex textual information related to telecommunications fraud.

\end{itemize}

\subsubsection{Question Answering}
Question answering tasks are a core component in NLP, aimed at developing systems to automatically answer questions about a given text. In the field of public security, question answering systems can assist law enforcement officers in quickly obtaining detailed information about specific cases, thereby enhancing work efficiency and decision-making quality. They can also address public inquiries regarding police services, legal regulations, and other related matters. For instance, through a question answering system, police can swiftly access vital details about specific cases, such as “What is the main evidence in a particular case?” or “What thefts have recently occurred in a certain area?” CPSDBench provides two datasets to test the capabilities of LLMs in Chinese public security question answering.

\begin{itemize}
\item[1)]
Police Case Reading Comprehension Dataset:  This dataset compiles reading comprehension questions based on actual police case files, covering both objective and subjective reading comprehension sections. For the objective part, LLMs are required to analyze the questions posed based on the provided case materials and offer answers in the form of "yes," "no," or "unknown." In the subjective reading comprehension section, LLMs need to synthesize information directly related to the question from the case materials, such as names of individuals, locations, times, and actions, to form appropriate responses. This dataset encompasses 70 objective questions, 70 subjective questions requiring high precision, and 60 subjective questions with relatively lower precision requirements.
\item[2)]
Police General Knowledge Test: The data is derived from the Chinese Police Law Enforcement Qualification Examination, aimed at assessing the LLMs legal knowledge and law enforcement capabilities related to policing, to determine their potential to assist police officers of various categories. It includes 70 true-false questions, 70 single-choice questions, and 60 multiple-choice questions.
\end{itemize}

\subsubsection{Text Generation}
The objective of text generation tasks is to automatically produce coherent and relevant textual content. In the field of public security, this can be utilized for the automatic generation of crime reports, police bulletins, and legal documents, among others. Effective text generation tools can significantly alleviate the paperwork burden of law enforcement officers, thereby enhancing work efficiency. For example, automatically generating detailed drafts of crime reports based on basic case information and evidence. CPSDBench provides two datasets for evaluating the application of LLMs in Chinese public security text generation.
\begin{itemize}
\item[1)]
Police Case Summary Dataset: This dataset focuses on generating concise and precise summaries of police cases, aimed at presenting key information in a succinct and clear manner. Its fundamental purpose is to evaluate the model's ability to effectively integrate and summarize essential case details.
\item[2)]
Police Cases Reports Generation: This task provides a range of basic information on public security cases, enabling LLMs to generate clear, accurate, and consistent comprehensive police reports based on this information. The core objective of this dataset is to conduct an in-depth evaluation of the quality of generation and readability of texts by LLMs in the context of producing public security reports, with the aim of enhancing the informational value of these reports. This, in turn, is expected to provide more effective support for the decision-making processes of law enforcement agencies and aid in the public's understanding and response.
\item[3)]
Police Enforcement Cases and Legal Analysis: This dataset collects public security cases from real-world scenarios, aiming to require LLMs to conduct in-depth analyses of the provided cases and related legal provisions, and on this basis, generate solutions for problem-solving. The purpose of designing this dataset is to assess the abilities of LLMs in understanding complex law enforcement scenarios, legal logic analysis, and applying legal knowledge to solve practical problems.
\end{itemize}

\subsection{Baselines and Prompt}
In the context of baseline models, a selection of ten representative LLMs proficient in processing Chinese-language tasks was made for the purpose of testing. These baselines encompass both commercialized LLMs and models that are either open-source or semi-open-source. Specifically, the LLMs subjected to evaluation include: ChatGPT, GPT-4, ChatGLM, Minimax-abab, Baichuan2, XVERSE, QWen, ERNIE Bot, and Atom (a model continuously trained in the Chinese domain based on LLaMA-2). The foundational attributes of each model are delineated in Table~\ref{tab:table2}.

\begin{figure}
  \centering
  \includegraphics[scale=0.6]{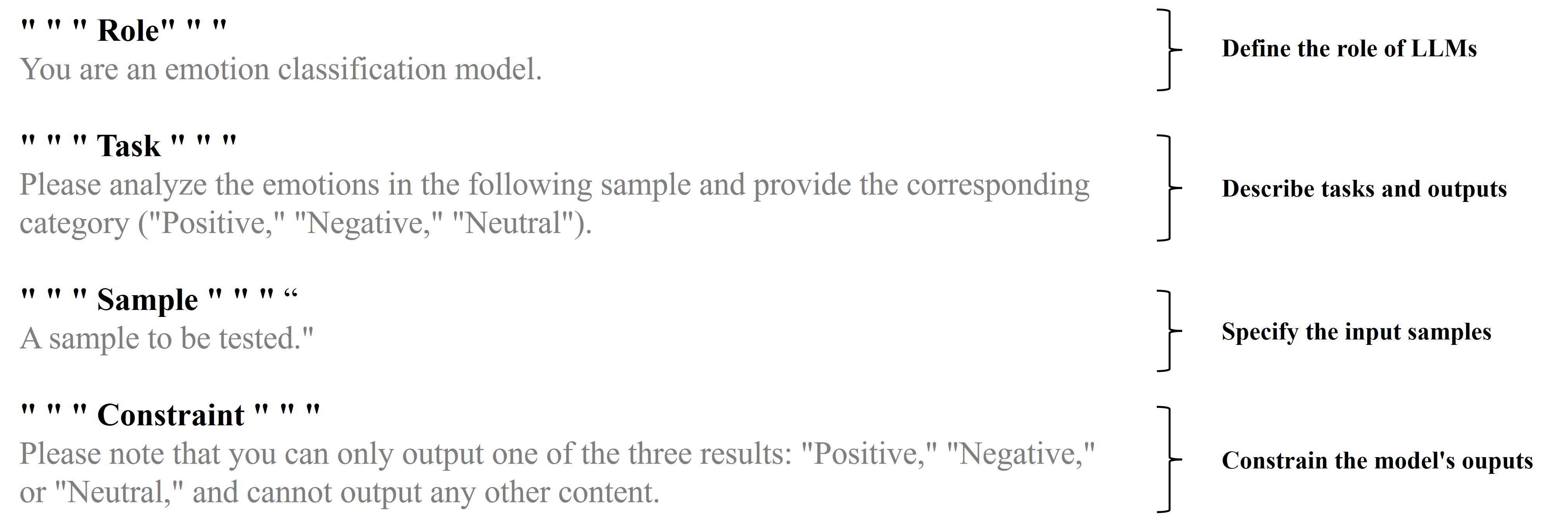}
  \caption{CPSDBench has designed appropriate prompts tailored to different tasks. Here, prompts for text classification and information extraction tasks are enumerated, primarily encompassing four key elements: role, task, input, and constraints.(Note that we used the Chinese version of the prompt during the evaluation process.)}
  \label{fig:fig2}
\end{figure}

Moreover, to accommodate the needs of different tasks in the public security domain, this study meticulously developed a specialized prompt engineering design framework. This framework comprises four critical components: role definition, task description, input specifications, and operational constraints. The intent of this design is to optimize the task processing capabilities of LLMs. By precisely delineating the nature of roles and tasks, clearly specifying the format and requirements of input data, and establishing specific constraints on operations, the framework aims to guide and control the model’s execution of specific public security tasks. This systematic approach to prompt engineering design not only enhances the model's efficiency and accuracy in executing particular tasks but also provides a vital reference framework for the deployment of models in complex application scenarios. The nuances of this prompt engineering process are detailed in Figure~\ref{fig:fig2}.

\begin{table}
 \caption{Table1: Details of baseline LLMs to be evaluated. It lists the model's name, parameter size (indicated as "$\sim$" to denote undisclosed specifics), URL, and access modes (API access or web-based access).}
  \centering
  \begin{tabular}{lllll}
    \toprule
    Model    & Organization & Size (\#P)   & URL & Access mode \\
    \midrule
    GPT-3.5-turbo & OpenAI & $\sim$ & \url{chat.openai.com}  & API \\
    GPT-4 & OpenAI & $\sim$ & \url{chat.openai.com}  & API \\
    Gemini-pro & Google & $\sim$ & \url{gemini.google.com}  & API \\
    ChatGLM-4 & THUDM \& Knowledge Atlas & 130B & \url{chatglm.cn}   & API \\
    MiniMax-ABAB-5.5 & MiniMax & $\sim$ & \url{api.minimax.chat}   & API \\
    Baichuan2\cite{yang2023baichuan} & Baichuan Inc. & $\sim$ & \url{www.baichuan-ai.com}   & API \\
    XVERSE & XVERSE & 13B,65B & \url{chat.xverse.cn}   & API    \\
    iFLYTEK Spark-3.0 & iFLYTEK CO.LTD. & $\sim$ & \url{xinghuo.xfyun.cn}   & API \\
    Qwen\cite{bai2023qwen} & Alibaba.com Corporation & 72B & \url{tongyi.aliyun.com/qianwen} & API \\
    ERNIE Bot\cite{sun2021ernie} & Baidu, Inc. & $\sim$ & \url{yiyan.baidu.com} & API \\
    SenseChat & SenseTime & $\sim$ & \url{chat.sensetime.com} & API \\
    Atom (LLaMA-2-Chinese) & Llama Family & 1B,7B & \url{llama.family} & API \\
    \bottomrule
  \end{tabular}
  \label{tab:table2}
\end{table}

\subsection{Evaluation Metrics}
Given that CPSDBench encompasses a variety of NLP tasks, we have selected different evaluation metrics based on the characteristics of each task.

For text classification tasks, we have chosen Accuracy, Precision, Recall, and F1-Score as evaluation metrics. The computation of this metric is illustrated in Equation (~\ref{eq:eq1}).
\begin{equation}\label{eq:eq1}
Accuracy = \frac{TP+TN}{TP+TN+FP+FN}
\end{equation}

For information extraction tasks, the typical metrics are Precision, Recall, and F1-Score. The computation of these metrics is illustrated in Equations (~\ref{eq:eq2})-(~\ref{eq:eq4}). However, these metrics often focus on calculating results that are exact matches. The output of LLMs may differ literally from the labels but be semantically equivalent. For instance, the label for a time entity in a sample might be '9:00 am on August 6, 2021', while the LLM's output could be 'around 9:00 am on August 6, 2021', where the additional word 'around' also originates from the input sample. From a human perspective, such LLM predictions should also be considered correct. Moreover, there are more complex scenarios in Chinese. In light of this, we have designed a hybrid evaluation metric. It comprises two steps: firstly, calculating the exact match score between the LLM's output and the label. Secondly, for predictions that are not exact matches, we calculate the Levenshtein distance, as shown in Equation (~\ref{eq:eq5}), and deem predictions above a certain threshold as correct for calculating a fuzzy match score. Finally, we obtain a comprehensive score by weighting these two types of scores, as depicted in Equation (~\ref{eq:eq6}). In the equation, Whereas, $a$ and $b$ represent two input strings, and $i$ and $j$ respectively denote the indices of string $a$ and string $b$. Additionally, for entities requiring high precision, such as the name and amount entities in telecommunication fraud case entity recognition, we have only calculated the exact match score, disregarding the fuzzy similar results.
\begin{equation}\label{eq:eq2}
Precision = \frac{TP}{TP+FP}
\end{equation}
\begin{equation}\label{eq:eq3}
Recall =  \frac{TP}{TP+FN}
\end{equation}
\begin{equation}\label{eq:eq4}
F1 = \frac{2 \times Precision \times Recall}{Precision + Recall}
\end{equation}
\begin{equation}\label{eq:eq5}
\operatorname{lev}_{a, b}(i, j)=\left\{\begin{array}{ll}{\max (i, j)} & {\text { if } \min (i, j)=0} \\ {\min \left\{\begin{array}{ll}{\operatorname{lev}_{a, b}(i-1, j)+1} \\ {\operatorname{lev}_{a, b}(i, j-1)+1} \\{\operatorname{lev}_{a, b}(i-1, j-1)+1_{\left(a_{i} \neq b_{j}\right)}} \end{array}\right.} & {\text { otherwise }}\end{array}\right.
\end{equation}

\begin{equation}\label{eq:eq6}
{Score}_{total} = \alpha \times {Score}_{exact} + (1 - \alpha) \times {Score}_{fuzzy}
\end{equation}

In the objective reading comprehension component of the question-answering task, this study adopts Accuracy as the primary evaluation metric to quantify the model's performance in identifying the correct answers. For generative question-answering and text generation tasks, the study selects Bilingual Evaluation Understudy (BLEU) and Recall-Oriented Understudy for Gisting Evaluation (ROUGE) as assessment metrics. These metrics are utilized to evaluate the model's performance in terms of the quality and accuracy of generated text. The specific computational processes are defined in Equations (~\ref{eq:eq7}) and (~\ref{eq:eq8}), respectively. Equation (~\ref{eq:eq7}) elaborates on the calculation method for the BLEU metric, focusing primarily on the lexical overlap between the generated text and the reference text. Equation (~\ref{eq:eq8}) describes the calculation process for the ROUGE metric, emphasizing the extent to which the generated text covers and retains the content of the original text. Together, these two metrics provide a comprehensive quantitative basis for evaluating the performance of models in generative question-answering and text generation tasks.

\begin{equation}\label{eq:eq7}
\operatorname{BLEU-N}(x)=b(x)\times\exp(\sum\limits_{N=1}^{N'}\alpha_N\log P_N)\quad
\end{equation}

\begin{equation}\label{eq:eq8}
\operatorname{ROUGE-N}(x)=\dfrac{\sum_{k=1}^n\sum_{w\in W}\min(c_w(x),c_w(r_k))}{\sum_{k=1}^n\sum_{w\in W}c_w(r_k)}
\end{equation}

\begin{equation}\label{eq:eq9}
Precision_{BERT} = \frac{1}{|C|} \sum_{c \in C} \max_{r \in R} \text{sim}(c, r)
\end{equation}

Furthermore, we have also selected the BERT Score\cite{zhang2019bertscore}, calculated based on the BERT-base model, as the evaluation criterion for subjective reading comprehension questions. This approach aims to measure the semantic similarity between LLMs' predictions and the labels from a semantic perspective.The computation of BERTScore involves three primary metrics: Precision, Recall, and the F1 Score, as illustrated in equation(~\ref{eq:eq9})-(~\ref{eq:eq11}). Here, $C$ represents the set of words in the candidate sentence, while $R$ denotes the set of words in the reference sentence. The term $sim(c,r)$ measures the similarity between the candidate word $c$ and the reference word $r$. Specifically, in subsequent experiments, we utilized the $F1_{BERT}$ as a evaluation metric for text generation tasks.

\begin{equation}\label{eq:eq10}
Recall_{BERT} =  \frac{1}{|R|} \sum_{r \in R} \max_{c \in C} \text{sim}(r, c)
\end{equation}

\begin{equation}\label{eq:eq11}
F1_{BERT} = \frac{2 \cdot Precision_{BERT} \cdot Recall_{BERT}}{Precision_{BERT} + Recall_{BERT}}
\end{equation}

\section{Results and Analysis}
\subsection{Performance Evaluation}
\begin{figure}
  \centering
  \includegraphics[scale=0.6]{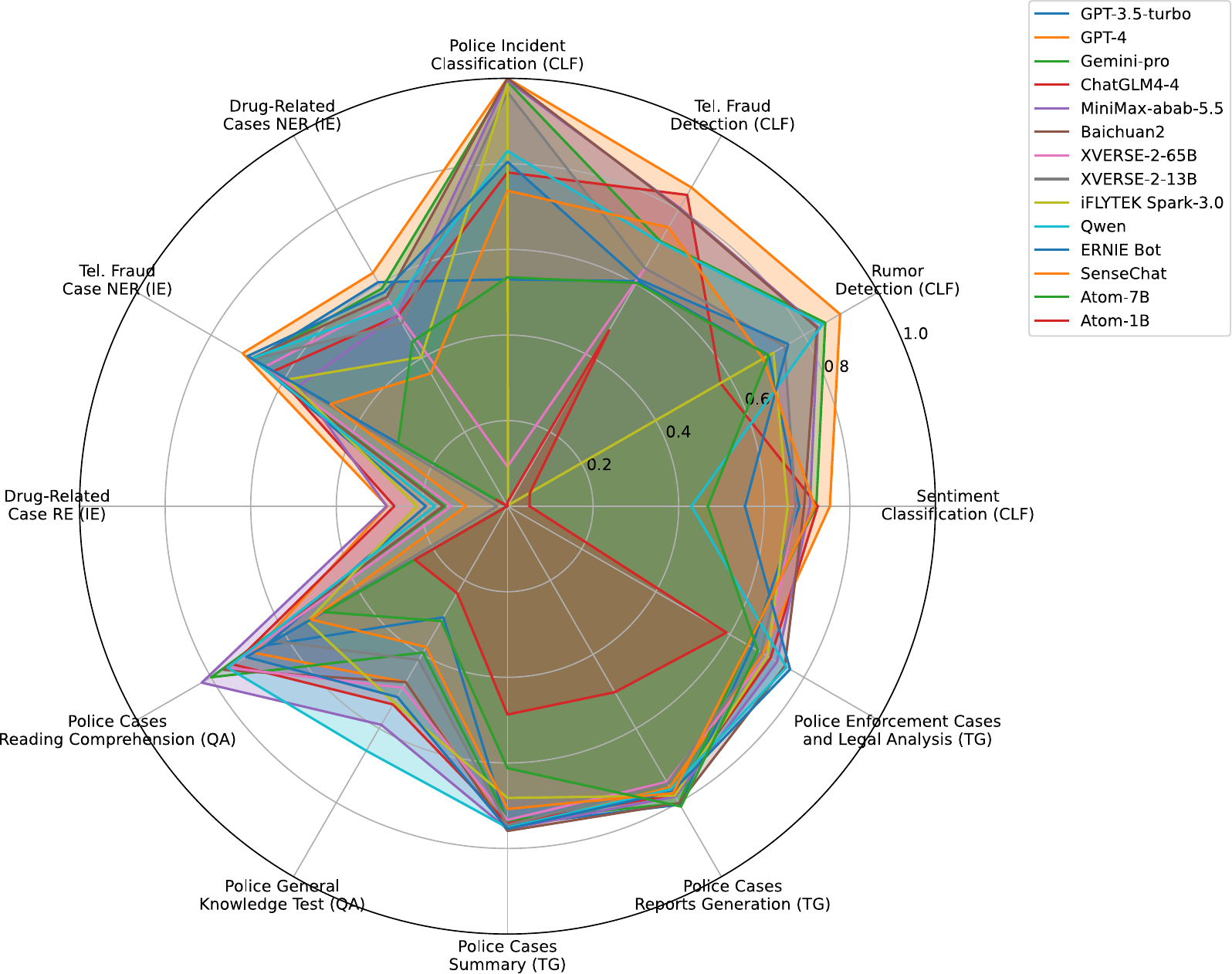}
  \caption{A comparative analysis of the overall performance of mainstream LLMs on CPSDBench.For information extraction, the F1 score is used as the final score. The text generation task employs the BERT score as its final score. CLF refers to text classification tasks, IE denotes relation extraction tasks, QA represents question-answering tasks, and TG signifies text generation tasks.}
  \label{fig:fig11}
\end{figure}

In this study, we conducted a comprehensive evaluation of advanced LLMs to explore their performance across different natural language processing tasks. The results are shown in Table~\ref{tab:table21}-Table~\ref{tab:table7}.Through systematic assessments in four key tasks: text classification, information extraction, question answering, and text generation, our aim was to unveil the strengths and limitations of mainstream LLMs in public security tasks.To achieve a more intuitive and comprehensive comparison of the performance of LLMs, this study also utilized radar charts to visually present the overall efficacy of the models, as shown in Figure~\ref{fig:fig11}. When invoking APIs to access LLMs, we ensure, as much as possible, the consistency of key parameters such as temperature, top-k, and top-p.

Firstly, in terms of overall performance, GPT-4 exhibited outstanding performance across all evaluation tasks, particularly in the natural language understanding tasks of text classification and information extraction, where its performance surpassed that of other models. This finding underscores GPT-4's efficient capability in understanding complex instructions and following specific guidelines.

This study further delves into the impact of language specificity on model performance. It particularly highlights that models focusing on Chinese, such as ChatGLM-4, Minimax-abab, and Baichuan2, exhibit superior performance in text generation and question-answering tasks compared to GPT-4. This finding underscores the distinct advantage of models pre-trained on rich Chinese corpora in handling tasks specific to the Chinese language. Such Chinese corpora enable LLMs to effectively learn key knowledge within the Chinese domain, such as public safety laws and law enforcement procedures, thereby proving the importance of language-specific pre-training in enhancing model performance in specific linguistic contexts.

The difficulty level of tasks also has a decisive impact on the performance of LLMs. For instance, in the task of relation extraction, models are first required to identify entities within the text and then analyze the relationships between these entities. This complexity makes such tasks significantly more challenging than text classification or entity recognition tasks, resulting in relatively lower scores for baseline models in these areas. At the same time, multiple-choice questions, compared to single-choice questions, offer a broader range of answer options, increasing the likelihood of LLMs missing or selecting too many options, thereby lowering their scores on such question types. Moreover, police enforcement cases and legal analysis tasks demand higher capabilities in text generation fluency, case logic reasoning, and mastery of public safety-related legal knowledge. The results indicate that only LLMs trained on Chinese-specific domain corpora achieve better scores in these tasks.

Our analysis also compared the performance disparities between proprietary and open-source models. Generally, proprietary models outperform open-source models. This can be attributed to proprietary models typically having a larger number of parameters and access to more extensive private training datasets, factors that collectively influence the final performance of the models.

Furthermore, we analyzed the impact of parameter scale on LLMs. By comparing the performance of different parameter scale versions of the Atom and XVerse models, we found that the number of parameters plays a key role in enhancing natural language understanding capabilities. However, this impact is less pronounced in text generation tasks. Surprisingly, the 13B version of XVerse even outperformed its 65B version in text generation tasks, suggesting that besides model size, other factors such as model architecture optimization, training strategies, or data quality may have a significant impact on the performance of LLMs.

Finally, input length is an important factor affecting the performance of LLMs in public security tasks. We observed the variation in F1 scores of LLMs when processing texts of different lengths in the task of extracting information from drug-related cases, as illustrated in Figure 1. The analysis indicates that for the majority of models, the predictive capability of LLMs tends to decline with an increase in the length of the input text.

\begin{table}
 \caption{The Results of LLMs in Text Classification Tasks.}
  \centering
  \begin{tabular}{lllll}
    \toprule
    Model     & Sentiment Classification   & Rumor Detection & Tel. Fraud Detection & Police Incident Classification \\
    \midrule
    GPT-3.5-turbo & 0.6817 & 0.7050  & 0.6075 & 0.5300 \\
    GPT-4 & 0.7533 & 0.8975  & 0.8600  & 1.000 \\
    Gemini-pro & 0.7217 & 0.8575  & 0.7150  & 0.9925 \\
    ChatGLM-4 & 0.7250 & 0.5750  & 0.8400  & 0.7800 \\
    MiniMax-ABAB-5.5 & 0.7067 & 0.8375   & 0.8050  & 0.9950 \\
    Baichuan2 & 0.6933 & 0.8350  & 0.8025  & 1.000 \\
    XVERSE-2-65B & 0.7100 & 0.8175  & 0.4925     & 0.0925 \\
    XVERSE-2-13B & 0.6733 & 0.7475 &  0.6425     & 0.9675 \\
    iFLYTEK Spark-3.0 & 0.6550 & 0.7175  & 0.0025  & 0.9875 \\
    Qwen & 0.4300 & 0.8500 & 0.7125  & 0.8300 \\
    ERNIE Bot & 0.5550 & 0.7575 & 0.6125  & 0.8050 \\
    SenseChat & 0.7183 & 0.6900  & 0.7525  & 0.7375 \\
    Atom-7B & 0.4683 & 0.7050 & 0.6025  & 0.5350 \\
    Atom-1B & 0.0516 & 0.0600 & 0.4750  & 0.010 \\
    \bottomrule
  \end{tabular}
  \label{tab:table21}
\end{table}

\begin{table}
 \caption{The Results of LLMs in Information Extraction Tasks.}
  \centering
  \begin{tabular}{llllllllll}
    \toprule
    \multirow{2}*{Model} & \multicolumn{3}{c}{Drug Case NER} & \multicolumn{3}{c}{Tel. Fraud Case NER} & \multicolumn{3}{c}{Drug Case RE}  \\
		 & Precision  & Recall & F1 Score  & Precision  & Recall & F1 Score  & Precision  & Recall & F1 Score \\
    \midrule
    GPT-3.5-turbo & 0.7072 & 0.5952  & 0.6045& 0.6967 &  0.6870 &  0.6845& 0.1777 &  0.2243 &  0.1907 \\
    GPT-4 & 0.6721 & 0.6352  & 0.6295& 0.7218 &  0.7224 &  0.7147& 0.2533 &  0.3401 &  0.2809 \\
    Gemini-pro & 0.6792 & 0.5715  & 0.5873&  0.6873 &  0.6952 &  0.6830 &  0.1534 &  0.1573 &  0.1456 \\
    ChatGLM-4 & 0.5939 & 0.5049  & 0.5136& 0.6347 &  0.6453 &  0.6310& 0.2216 &  0.3548 &  0.2643 \\
    MiniMax-ABAB-5.5 & 0.6108 & 0.5078   &  0.5159& 0.5805 &  0.5623 &  0.5626& 0.2534 &  0.3610 &  0.2823 \\
    Baichuan2 & 0.6566 & 0.5556  & 0.5644& 0.7042 &  0.6996 &  0.6909& 0.1344 &  0.1896 &  0.1507 \\
    XVERSE-2-65B & 0.6778 & 0.5198  & 0.5526& 0.7149 &  0.6281 &  0.6530 &0.1188 &  0.1642 &  0.1309  \\
    XVERSE-2-13B & 0.5937 & 0.4687   & 0.4942 & 0.7136 &  0.6806 &  0.6872  & 0.0233 &  0.0292 &  0.0248  \\
    iFLYTEK Spark-3.0 & 0.4612 & 0.3913  & 0.4025& 0.6053 &  0.5923 &  0.5925 & 0.1655 &  0.3210 &  0.2117\\
    Qwen & 0.5729 & 0.5654 & 0.5394& 0.6988 &  0.7009 &  0.6905  & 0.1638 &  0.1933 &  0.1703\\
    ERNIE Bot & 0.6507 & 0.5751 & 0.5773 & 0.7050 &  0.7127 &  0.7014 &0.0017 &  0.0017 &  0.0017\\
    SenseChat & 0.4211 & 0.3553  & 0.3587& 0.5184 &  0.4644 &  0.4776 & 0.0764 &  0.1482 &  0.0973\\
    Atom-7B & 0.6385 & 0.3905 & 0.4441& 0.3564 &  0.2698 &  0.2950 & 0.0000 &  0.0000 &  0.0000\\
    Atom-1B & 0.0007  & 0.0014 & 0.0009 & 0.0396 &  0.0285 &  0.0312 & 0.0000 &  0.0000 &  0.0000\\
    \bottomrule
  \end{tabular}
  \label{tab:table3}
\end{table}

\begin{table}
 \caption{The Results of LLMs in Question Answering Tasks. (1)"T/F/U" and "T/F" respectively stand for True-False-Unknown questions and True-False questions. (2)"Exactitude" refers to those categories of questions that demand higher precision in the outcomes, primarily revolving around queries related to personal names. In assessing the answers to these types of questions, more stringent evaluation criteria are employed. (3)"Fuzzy" denotes other types of question and answer formats, encompassing queries related to locations, times, and methods. These queries utilize more lenient evaluation criteria (such as fuzzy matching) when evaluating responses. (4)"SCQ" and "MCQ" respectively stand for Single Choice Questions and Multiple Choice Questions. (5)"Total" is the weighted score obtained according to the proportion of questions, representing the model's overall effectiveness.}
  \centering
  \begin{tabular}{lllllllll}
    \toprule
    \multirow{2}*{Model} & \multicolumn{4}{c}{Police Cases Reading Comprehension} & \multicolumn{4}{c}{Police General Knowledge Test}  \\
		 & Total & T/F/U  & Exactitude & Fuzzy & Total & T/F & SCQ & MCQ \\
    \midrule
    GPT-3.5-turbo &  0.6500 & 0.7000  &0.4143 &0.8667 & 0.3000 &  0.4429 &  0.3857 &  0.0333\\
    GPT-4 & 0.6850 & 0.9143  & 0.3571 & 0.8000  & 0.4750 &  0.5143 &  0.6143 &  0.2667\\
    Gemini-pro &  0.8000 & 0.8286 & 0.7000 & 0.8833  & 0.3950 &  0.4714 &  0.5286 &  0.1500\\
    ChatGLM-4 & 0.7400 & 0.8714  & 0.4714 & 0.9000  & 0.5350 &  0.3000 &  0.8429 &  0.4500\\
    MiniMax-ABAB-5.5 & 0.8250 &  0.8714   &   0.7429 & 0.8667 & 0.5900 &  0.6286 &  0.7571 &  0.3500\\
    Baichuan2 & 0.7650 & 0.7857  &0.6857 & 0.8333 & 0.4750 &  0.6714 &  0.5143 &  0.2000\\
    XVERSE-2-65B & 0.7550 & 0.8143  & 0.5857 & 0.8833    & 0.4900 &  0.6143 &  0.5714 &  0.2500\\
    XVERSE-2-13B & 0.6300 & 0.7429   &  0.3000 &  0.8833  & 0.4150 &  0.4857 &  0.4857 &  0.2500\\
    iFLYTEK Spark-3.0 &  0.5400 & 0.7000  & 0.2714& 0.6667 & 0.5300 &  0.5429 &  0.6429 &  0.3833 \\
    Qwen &0.7550 & 0.7429 &0.7286 & 0.8000 & 0.6600 &  0.5571 &  0.8571 &  0.5500\\
    ERNIE Bot & 0.7050 & 0.8714 &  0.4000 & 0.8667 &0.5150 &  0.2714 &  0.7714 &  0.5000 \\
    SenseChat & 0.5300 & 0.7286  & 0.1429 & 0.7500  & 0.3800 &  0.3571 &  0.5857 &  0.1667\\
    Atom-7B &  0.4950 & 0.4429 & 0.3143 & 0.7667  & 0.3100 &  0.4286 &  0.2857 &  0.2000\\
    Atom-1B & 0.2500   & 0.4143 & 0.0714 & 0.2667  & 0.2350 &  0.3857 &  0.2714 &  0.0167\\
    \bottomrule
  \end{tabular}
  \label{tab:table4}
\end{table}

\begin{table}
 \caption{The Results of LLMs in the Police Cases Summary Task.}
  \centering
  \begin{tabular}{llllllll}
    \toprule
    Model  & BERTScore & C-BLEU  & W-BLEU & C-ROUGE-1 & C-ROUGE-L & W-ROUGE-1 & W-ROUGE-L\\
    \midrule
    GPT-3.5-turbo & 0.7463 & 0.1925  & 0.1076 & 0.5248  & 0.4110 & 0.4215  & 0.3416 \\
    GPT-4 & 0.7545 & 0.2333  & 0.1430 & 0.5592  & 0.4588 & 0.4464  & 0.3756 \\
    Gemini-pro & 0.7340 & 0.1686  & 0.0897 & 0.5012  & 0.3895 & 0.4154  & 0.3349 \\
    ChatGLM-4 & 0.7415 &0.1949  & 0.0987 & 0.5364  &0.4144 & 0.4104 & 0.3261 \\
    MiniMax-ABAB-5.5 & 0.7549 & 0.2287  & 0.1601 & 0.5499  &0.4659 & 0.4489  & 0.3906 \\
    Baichuan2& 0.7591 & 0.2304  & 0.1449 &  0.5597  &0.4594 & 0.4456  & 0.3770 \\
    XVERSE-2-65B & 0.7332 &0.1681  & 0.0840 & 0.4866  & 0.3825 & 0.3704  & 0.3034 \\
    XVERSE-2-13B & 0.7453 & 0.2128  & 0.1279 & 0.5240  & 0.4213 & 0.4140  & 0.3429 \\
    iFLYTEK Spark-3.0 & 0.6820 & 0.1704  & 0.1050 & 0.4419  & 0.3515 & 0.3488  & 0.2908 \\
    Qwen & 0.7514 &0.2482  & 0.1771 & 0.5446  &0.4602 & 0.4497  & 0.3904 \\
    ERNIE Bot & 0.7542 & 0.2352  & 0.1468 &0.5523  & 0.4489 & 0.4505  & 0.3756\\
    SenseChat & 0.7078 & 0.1309  & 0.0789 & 0.3428  & 0.2708 & 0.2774  & 0.2261 \\
    Atom-7B & 0.6123 & 0.1177  & 0.0707 &0.3823 &  0.3204 & 0.3000  &  0.2611 \\
    Atom-1B & 0.4870 & 0.0205  & 0.0168 & 0.0739  & 0.0621 & 0.0462 & 0.0417 \\
    \bottomrule
  \end{tabular}
  \label{tab:table5}
\end{table}

\begin{table}
 \caption{The Results of LLMs in the Police Cases Reports Generation Task.}
  \centering
  \begin{tabular}{llllllll}
    \toprule
    Model  & BERTScore & C-BLEU  & W-BLEU & C-ROUGE-1 & C-ROUGE-L & W-ROUGE-1 & W-ROUGE-L\\
    \midrule
    GPT-3.5-turbo & 0.8051 &  0.3103 &  0.2040 &  0.7046 &  0.6022 &  0.6257 &  0.5386 \\
    GPT-4 & 0.7539 &  0.2525 &  0.1546 &  0.6203 &  0.4882 &  0.4881 &  0.4019 \\
    Gemini-pro &0.7962 &  0.3298 &  0.2158 &  0.6896 &  0.5717 &  0.5917 &  0.4960 \\
    ChatGLM-4 & 0.7777 &  0.3113 &  0.2007 &  0.6638 &  0.5393 &  0.5461 &  0.4512 \\
    MiniMax-ABAB-5.5 &0.7871 &  0.3216 &  0.2146 &  0.6789 &  0.5704 &  0.5822 &  0.4986 \\
    Baichuan2& 0.7965 &  0.3415 &  0.2260 &  0.6942 &  0.5732 &  0.5932 &  0.4949 \\
    XVERSE-2-65B &0.7393 &  0.2082 &  0.1144 &  0.5605 &  0.4142 &  0.4165 &  0.3237 \\
    XVERSE-2-13B & 0.7472 &  0.2368 &  0.1371 &  0.5864 &  0.4476 &  0.4516 &  0.3610 \\
    iFLYTEK Spark-3.0 & 0.7794 &  0.3090 &  0.2063 &  0.6560 &  0.5395 &  0.5521 &  0.4670 \\
    Qwen &0.7597 &  0.2581 &  0.1551 &  0.6102 &  0.4750 &  0.4837 &  0.3866 \\
    ERNIE Bot & 0.7674 &  0.2776 &  0.1778 &  0.6312 &  0.5079 &  0.5132 &  0.4202\\
    SenseChat & 0.7359 &  0.2536 &  0.1593 &  0.5788 &  0.4635 &  0.4717 &  0.3901 \\
    Atom-7B & 0.7253 &  0.2369 &  0.1586 &  0.5489 &  0.4690 &  0.4966 &  0.4304 \\
    Atom-1B & 0.4896 &  0.0287 &  0.0183 &  0.1018 &  0.0658 &  0.0821 &  0.0582 \\
    \bottomrule
  \end{tabular}
  \label{tab:table6}
\end{table}

\begin{table}
 \caption{The Results of LLMs in the Police Enforcement Cases and Legal Analysis Task.}
  \centering
  \begin{tabular}{llllllll}
    \toprule
    Model  & BERTScore & C-BLEU  & W-BLEU & C-ROUGE-1 & C-ROUGE-L & W-ROUGE-1 & W-ROUGE-L\\
    \midrule
    GPT-3.5-turbo & 0.6588 &  0.0492 &  0.0200 &  0.3458 &  0.2496 &  0.2433 &  0.1909 \\
    GPT-4 & 0.6890 &  0.0973 &  0.0441 &  0.4181 &  0.2929 &  0.2954 &  0.2155 \\
    Gemini-pro & 0.6469 &  0.0620 &  0.0335 &  0.3150 &  0.2343 &  0.2397 &  0.1908 \\
    ChatGLM-4 & 0.7098 &  0.1425 &  0.0752 &  0.4601 &  0.3380 &  0.3535 &  0.2710 \\
    MiniMax-ABAB-5.5 &0.7275 &  0.1938 &  0.1252 &  0.4946 &  0.3985 &  0.3948 &  0.3285 \\
    Baichuan2& 0.7482 &  0.2394 &  0.1665 &  0.5362 &  0.4452 &  0.4377 &  0.3759 \\
    XVERSE-2-65B & 0.6998 &  0.1541 &  0.0875 &  0.4455 &  0.3267 &  0.3296 &  0.2597 \\
    XVERSE-2-13B & 0.6841 &  0.1028 &  0.0530 &  0.4061 &  0.2901 &  0.2946 &  0.2216\\
    iFLYTEK Spark-3.0 & 0.7016 &  0.1331 &  0.0794 &  0.4429 &  0.3309 &  0.3379 &  0.2655 \\
    Qwen & 0.7529 &  0.2542 &  0.1917 &  0.5343 &  0.4453 &  0.4533 &  0.3977 \\
    ERNIE Bot & 0.7626 &  0.2660 &  0.2023 &  0.5505 &  0.4565 &  0.4699 &  0.4026\\
    SenseChat & 0.6491 &  0.1275 &  0.0827 &  0.3329 &  0.2578 &  0.2656 &  0.2157 \\
    Atom-7B & 0.6756 &  0.0823 &  0.0481 &  0.3781 &  0.2918 &  0.2871 &  0.2250 \\
    Atom-1B & 0.5910 &  0.0212 &  0.0107 &  0.1941 &  0.1327 &  0.1489 &  0.1147 \\
    \bottomrule
  \end{tabular}
  \label{tab:table7}
\end{table}

\begin{figure}
  \centering
  \includegraphics[scale=0.39]{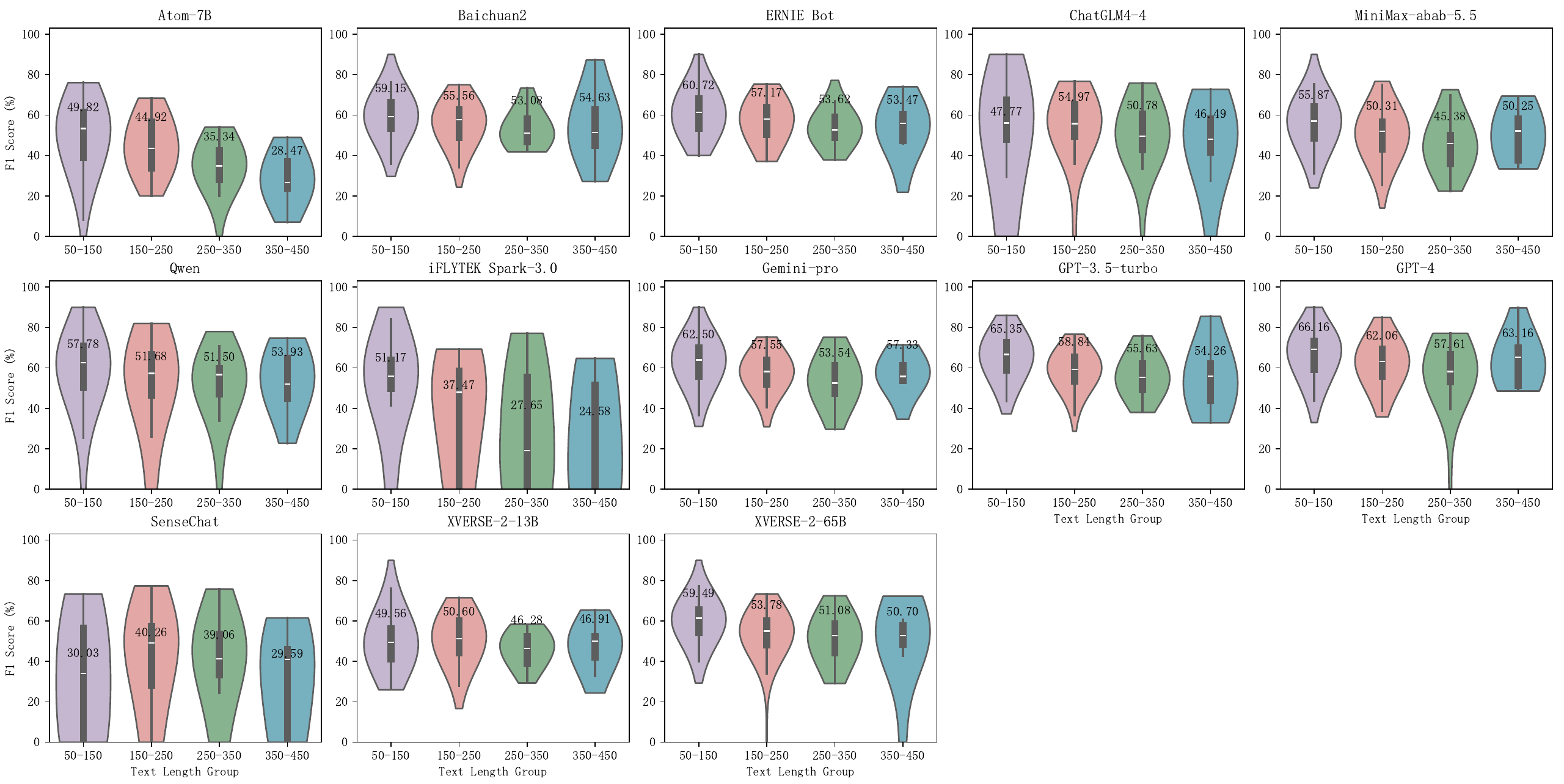}
  \caption{The impact of text length on the F1 scores for LLMs. To avoid the occurrence of negative values in the violin plot due to the kernel density estimation process, a phased processing approach was adopted for visualization. Due to the poor performance of the Atom-1B model in information extraction tasks, it is not shown in the figure.}
  \label{fig:fig10}
\end{figure}

\subsection{Error Analysis}
\begin{figure}
  \centering
  \includegraphics[scale=0.35]{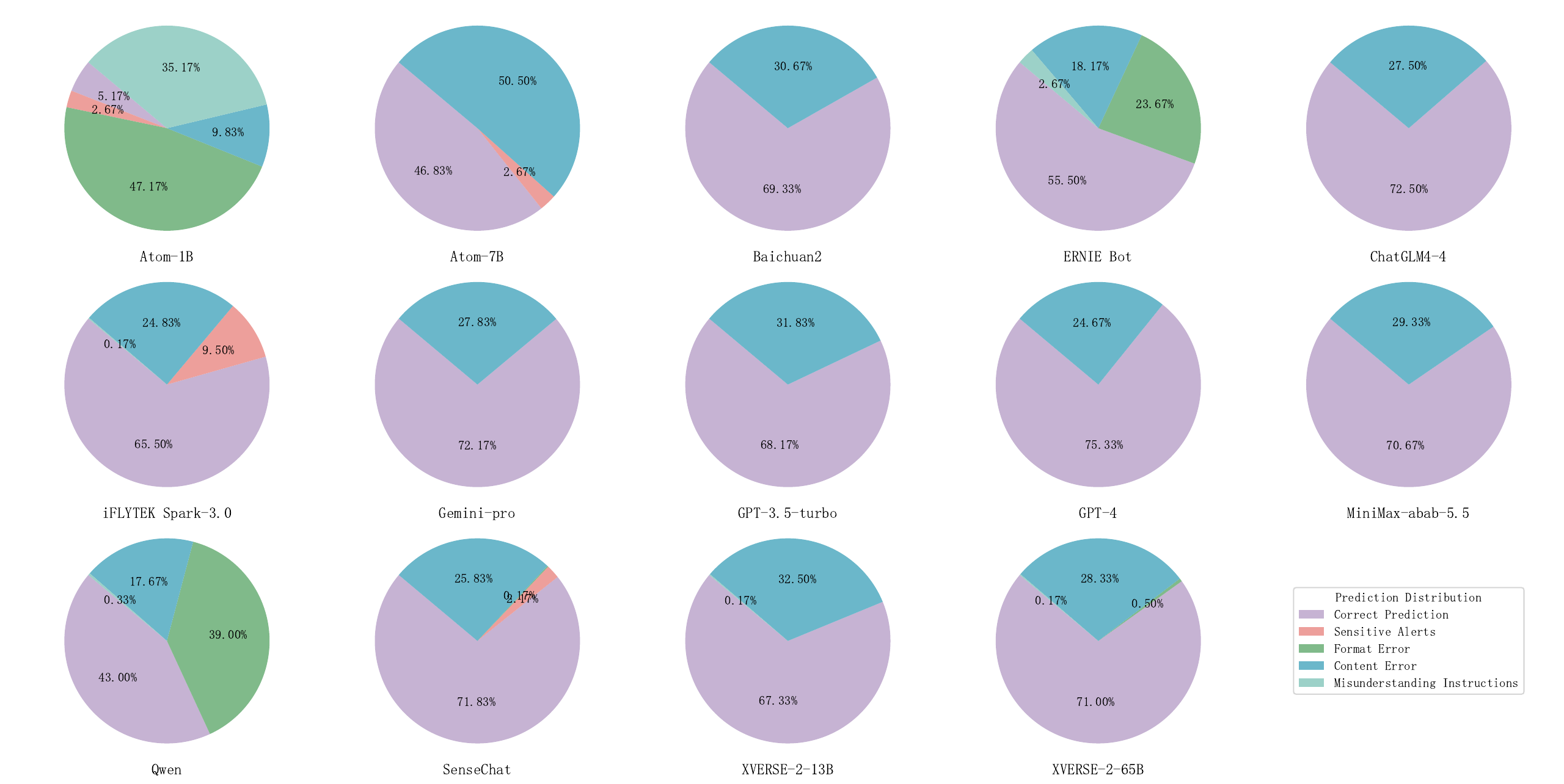}
  \caption{Error Type Distribution of LLMs in Sentiment Classification Task.}
  \label{fig:fig3}
\end{figure}

\begin{figure}
  \centering
  \includegraphics[scale=0.35]{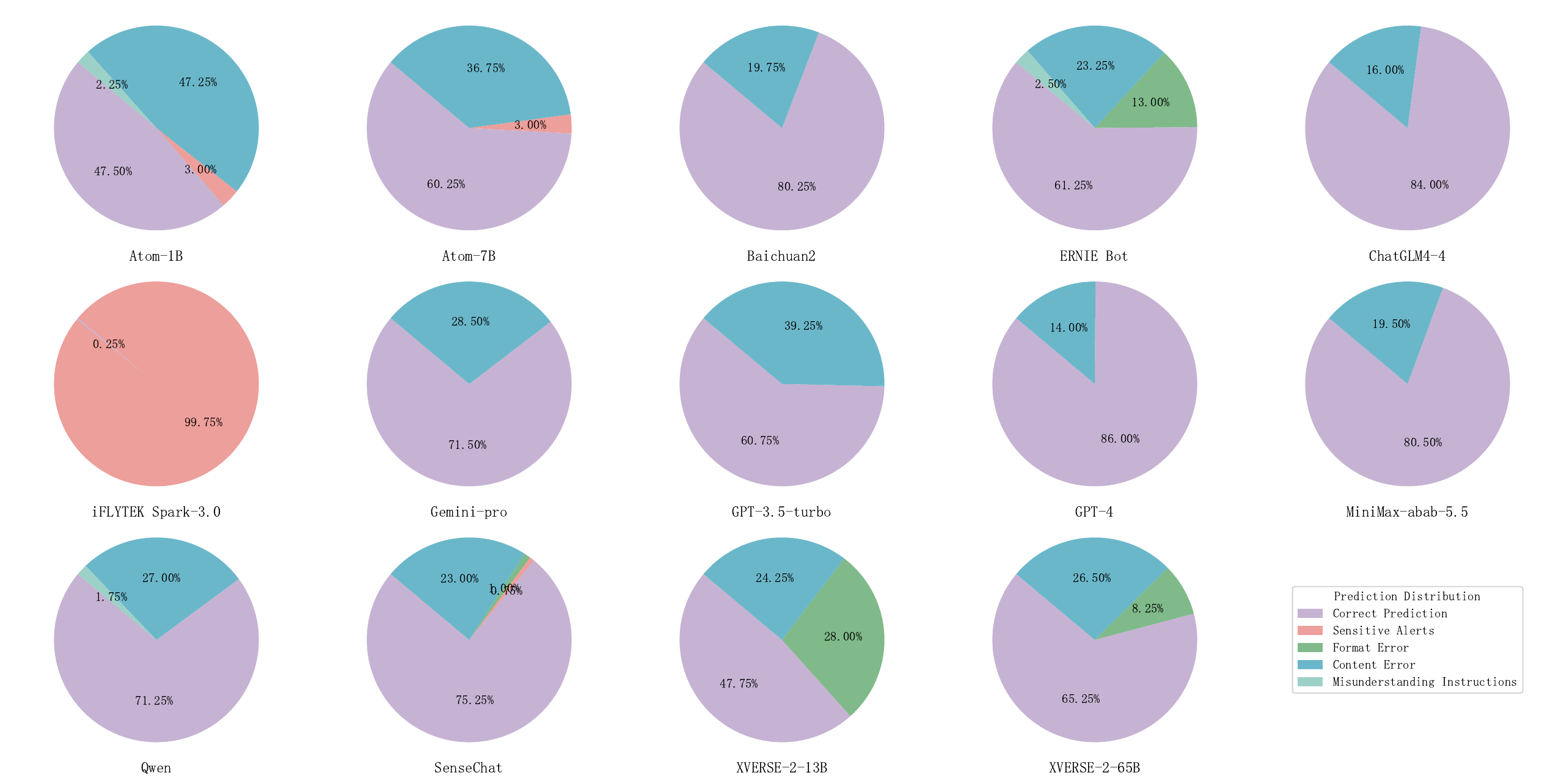}
  \caption{Error Type Distribution of LLMs in Telecommunications Fraud Detection Task.}
  \label{fig:fig4}
\end{figure}

To gain a deeper understanding of the limitations of existing LLMs in handling complex tasks, we conducted a comprehensive error analysis on examples that performed poorly in the CPSDBench tasks. These errors can primarily be categorized into four types: triggering sensitive alerts, incorrect output formatting, failure to understand instructions, and erroneous content generation. These categories reflect the challenges LLMs face in comprehension, generation, and safety aspects. Figure~\ref{fig:fig3} and figure~\ref{fig:fig4} illustrate the distribution of error types in LLMs across two representative tasks: sentiment classification and telecommunications fraud detection.

\subsubsection{Triggering Sensitive Alerts}
We observed that LLMs tend to trigger safety mechanisms when processing tasks containing sensitive content. Taking iFLYTEK Spark as an example, its built-in stringent user input review mechanism resulted in the model nearly refusing to answer all questions in tasks such as telecommunication fraud detection. This excessive cautiousness led to extremely low accuracy rates in such tasks, for instance, only a 2.5\% accuracy rate in the telecommunication fraud detection task. This indicates that LLMs might experience performance degradation in tasks involving sensitive information due to overly stringent safety filtering.

\subsubsection{Incorrect Output Format}
\begin{figure}
  \centering
  \includegraphics[width=\textwidth]{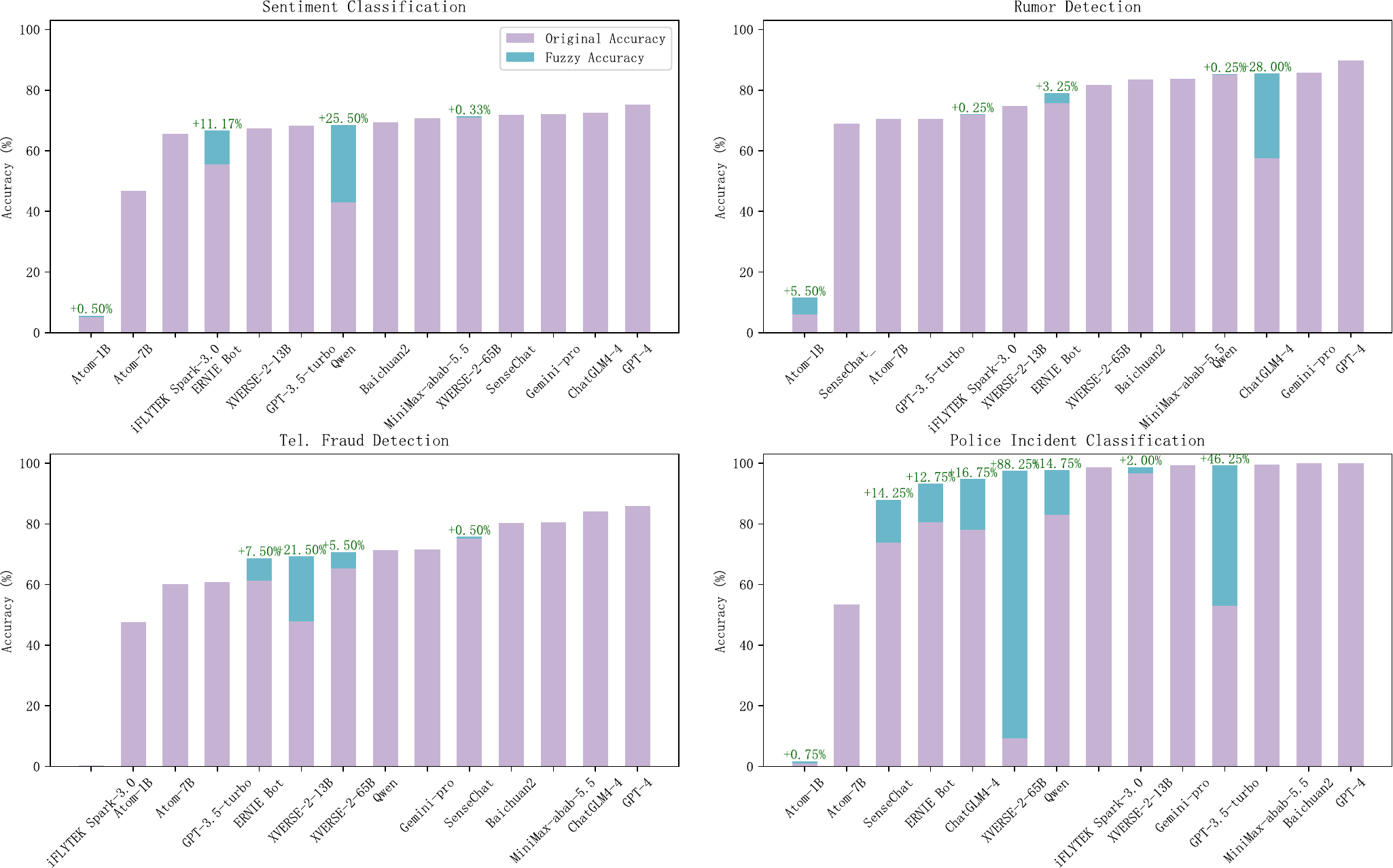}
  \caption{After adopting fuzzy accuracy as the evaluation metric, the performance of LLMs in text classification tasks changed. The accuracy improvement of models such as ChatGLM, Gemini, and XVERSE even exceeded 20\%.}
  \label{fig:fig5}
\end{figure}

Issues with output format represent one of the common error types for LLMs. In text classification and information extraction tasks, incorrect output formats directly impact the calculation of evaluation metrics. For example, the ChatGLM and ERNIE models incorporated additional explanatory information in their outputs, which did not comply with the task requirements for concise output formats. Moreover, when faced with open-ended questions, models tend to generate complete sentences rather than the required keywords or brief answers, leading to discrepancies between the predicted results and the standard answers.
To assess the actual performance of LLMs after disregarding formatting errors, this study implemented an accuracy calculation method based on the principles of fuzzy matching. Through this method, we observed significant variations in the accuracy of multiple LLMs in text classification tasks, with specific changes presented in Figure~\ref{fig:fig5}. Notably, the accuracy of some models improved by more than 20\% after adopting fuzzy matching for calculation. This demonstrates the significant importance of enhancing the formatting output capabilities of LLMs for tasks related to public security.

\subsubsection{Inability to Understand Instructions}
We discovered that models with smaller parameter sizes, such as Atom-1B, often struggle to accurately comprehend user instructions. These models may merely repeat the content of the prompt in text generation tasks rather than generating new, relevant information. This indicates that a model's comprehension and generative abilities are closely related to its parameter scale, with smaller models facing limitations in understanding complex instructions or generating pertinent content.

\subsubsection{Errors in Output Content}
\begin{figure}
  \centering
  \includegraphics[width=\textwidth]{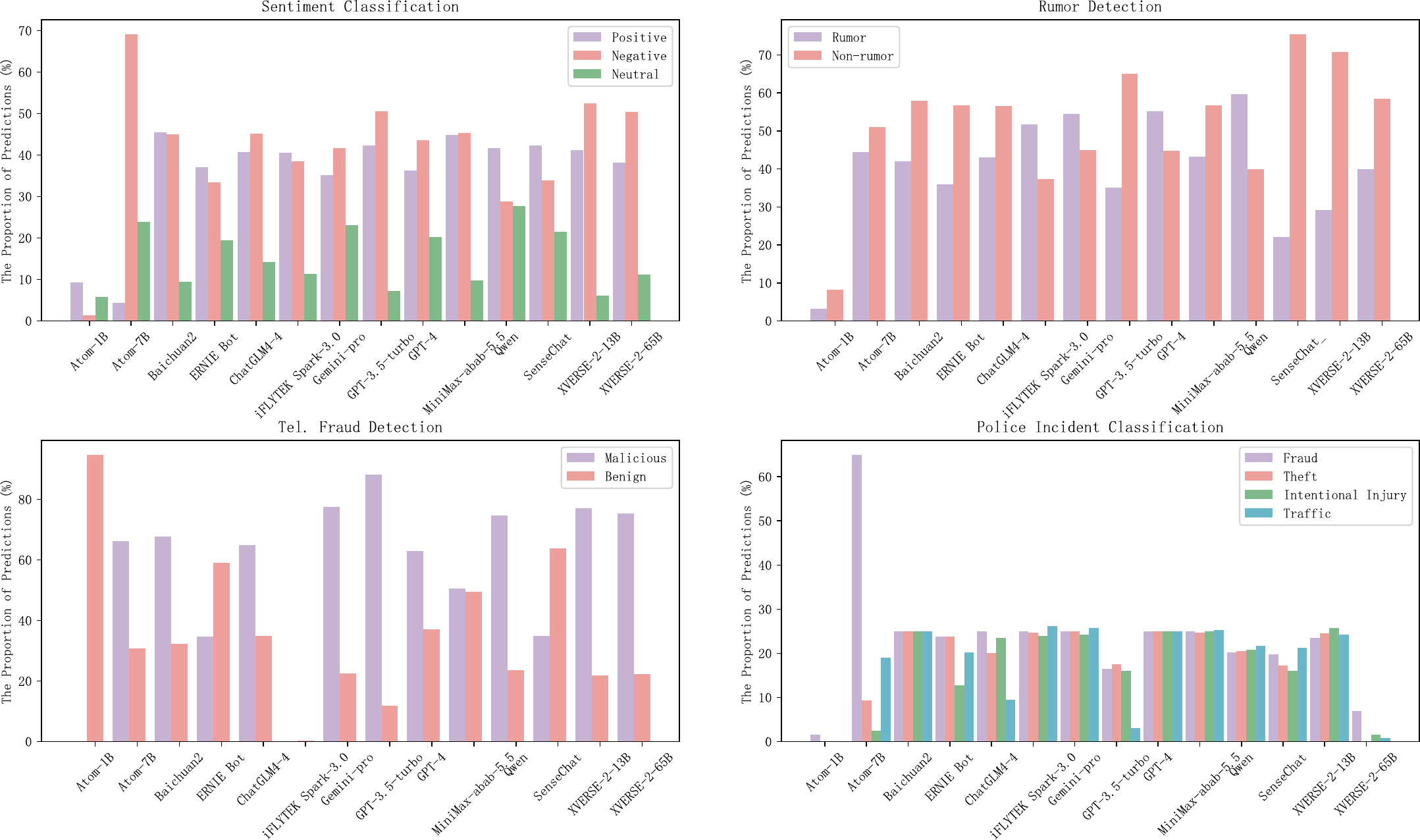}
  \caption{Predictive bias of LLMs in four text  classification tasks. (1) In the execution of sentiment classification tasks, the majority of models more frequently predict emotional tendencies as either "positive" or "negative," rather than opting for a "neutral" label. (2) In telecommunications fraud detection tasks, LLMs exhibit a tendency to output "malicious" labels, which may lead to a higher false positive rate when applying them to this task.}
  \label{fig:fig6}
\end{figure}

Regarding errors in output content, especially in information extraction tasks, some models exhibited issues with missing key entities. Moreover, in text classification tasks, some models like Atom-1B tended to excessively output labels of a specific category, such as classifying instances as "normal" in telecommunication fraud detection tasks. This may reflect biases in the model's training or overfitting to the training data distribution.Figures ~\ref{fig:fig6} illustrates the output bias of LLMs in three four classification tasks.

In summary, although LLMs have demonstrated significant potential for application in the public security domain, our analysis also explicitly points out various challenges they face in actual deployment. To overcome these challenges, future work needs to focus on improving the balance between model safety and usability, enhancing the flexibility and adaptability of output formats, improving model comprehension abilities, and optimizing the accuracy of content generation.
\section{Conclusion}
This study introduces CPSDBench, a benchmark specifically designed for evaluating LLMs in the Chinese public security domain. By collecting datasets from real-world public security scenarios and optimizing relevant evaluation metrics to better align with specific tasks, this benchmark conducts a comprehensive assessment of current LLMs across four dimensions: text classification, information extraction, question answering, and text generation. The findings indicate that LLMs, exemplified by ChatGPT, perform well in accomplishing most public security tasks, yet exhibit limitations in processing sensitive data and adversarial samples. This research not only aids in a deeper understanding and evaluation of existing models’ strengths and limitations in handling public security-related tasks but also provides significant references and insights for the future development of more suitable customized models for public security applications.

In our future research endeavors, we plan to undertake a series of measures to further enhance and deepen the evaluation of LLMs' application effectiveness in the public security domain. Firstly, we will dedicate efforts to collecting and integrating a broader and more diverse array of public security-related datasets. By covering a richer variety of scenarios and situations, we aim to examine the capabilities of LLMs in addressing complex, variable, and urgent public security issues. Moreover, we also plan to conduct in-depth optimization and expansion of the current evaluation metric system. This includes not only fine-tuning existing metrics to improve the precision and reliability of evaluations but also developing new evaluation metrics to better capture the nuanced differences and specific capabilities of LLMs in particular public security tasks. Through these efforts, we hope to provide more profound insights and guidance for the research and application of LLMs within the public security domain, thereby advancing the technological progress and development in this field.
\section*{Acknowledgments}
This study was supported in part by National Social Science Foundation Key Project of China (No.20AZD114).

\bibliographystyle{unsrt}  
\bibliography{CPSDBench}

\begin{thebibliography}{10}

\bibitem{touvron2023llama}
Hugo Touvron, Thibaut Lavril, Gautier Izacard, Xavier Martinet, Marie-Anne Lachaux, Timoth{\'e}e Lacroix, Baptiste Rozi{\`e}re, Naman Goyal, Eric Hambro, Faisal Azhar, et~al.
\newblock Llama: Open and efficient foundation language models.
\newblock {\em arXiv preprint arXiv:2302.13971}, 2023.

\bibitem{touvron2023llama2}
Hugo Touvron, Louis Martin, Kevin Stone, Peter Albert, Amjad Almahairi, Yasmine Babaei, Nikolay Bashlykov, Soumya Batra, Prajjwal Bhargava, Shruti Bhosale, Dan Bikel, Lukas Blecher, Cristian~Canton Ferrer, Moya Chen, Guillem Cucurull, David Esiobu, Jude Fernandes, Jeremy Fu, Wenyin Fu, Brian Fuller, Cynthia Gao, Vedanuj Goswami, Naman Goyal, Anthony Hartshorn, Saghar Hosseini, Rui Hou, Hakan Inan, Marcin Kardas, Viktor Kerkez, Madian Khabsa, Isabel Kloumann, Artem Korenev, Punit~Singh Koura, Marie-Anne Lachaux, Thibaut Lavril, Jenya Lee, Diana Liskovich, Yinghai Lu, Yuning Mao, Xavier Martinet, Todor Mihaylov, Pushkar Mishra, Igor Molybog, Yixin Nie, Andrew Poulton, Jeremy Reizenstein, Rashi Rungta, Kalyan Saladi, Alan Schelten, Ruan Silva, Eric~Michael Smith, Ranjan Subramanian, Xiaoqing~Ellen Tan, Binh Tang, Ross Taylor, Adina Williams, Jian~Xiang Kuan, Puxin Xu, Zheng Yan, Iliyan Zarov, Yuchen Zhang, Angela Fan, Melanie Kambadur, Sharan Narang, Aurelien Rodriguez, Robert Stojnic, Sergey Edunov, and Thomas
  Scialom.
\newblock Llama 2: Open foundation and fine-tuned chat models, 2023.

\bibitem{radford2018improving}
Alec Radford, Karthik Narasimhan, Tim Salimans, Ilya Sutskever, et~al.
\newblock Improving language understanding by generative pre-training.

\bibitem{schulman2017proximal}
John Schulman, Filip Wolski, Prafulla Dhariwal, Alec Radford, and Oleg Klimov.
\newblock Proximal policy optimization algorithms.
\newblock {\em arXiv preprint arXiv:1707.06347}, 2017.

\bibitem{ouyang2022training}
Long Ouyang, Jeffrey Wu, Xu~Jiang, Diogo Almeida, Carroll Wainwright, Pamela Mishkin, Chong Zhang, Sandhini Agarwal, Katarina Slama, Alex Ray, et~al.
\newblock Training language models to follow instructions with human feedback.
\newblock {\em Advances in Neural Information Processing Systems}, 35:27730--27744, 2022.

\bibitem{wei2022emergent}
Jason Wei, Yi~Tay, Rishi Bommasani, Colin Raffel, Barret Zoph, Sebastian Borgeaud, Dani Yogatama, Maarten Bosma, Denny Zhou, Donald Metzler, et~al.
\newblock Emergent abilities of large language models.
\newblock {\em arXiv preprint arXiv:2206.07682}, 2022.

\bibitem{brown2020language}
Tom Brown, Benjamin Mann, Nick Ryder, Melanie Subbiah, Jared~D Kaplan, Prafulla Dhariwal, Arvind Neelakantan, Pranav Shyam, Girish Sastry, Amanda Askell, et~al.
\newblock Language models are few-shot learners.
\newblock {\em Advances in neural information processing systems}, 33:1877--1901, 2020.

\bibitem{wei2022chain}
Jason Wei, Xuezhi Wang, Dale Schuurmans, Maarten Bosma, Fei Xia, Ed~Chi, Quoc~V Le, Denny Zhou, et~al.
\newblock Chain-of-thought prompting elicits reasoning in large language models.
\newblock {\em Advances in Neural Information Processing Systems}, 35:24824--24837, 2022.

\bibitem{du2022glm}
Zhengxiao Du, Yujie Qian, Xiao Liu, Ming Ding, Jiezhong Qiu, Zhilin Yang, and Jie Tang.
\newblock Glm: General language model pretraining with autoregressive blank infilling.
\newblock In {\em Proceedings of the 60th Annual Meeting of the Association for Computational Linguistics (Volume 1: Long Papers)}, pages 320--335, 2022.

\bibitem{zeng2022glm}
Aohan Zeng, Xiao Liu, Zhengxiao Du, Zihan Wang, Hanyu Lai, Ming Ding, Zhuoyi Yang, Yifan Xu, Wendi Zheng, Xiao Xia, et~al.
\newblock Glm-130b: An open bilingual pre-trained model.
\newblock {\em arXiv preprint arXiv:2210.02414}, 2022.

\bibitem{wu2023bloomberggpt}
Shijie Wu, Ozan Irsoy, Steven Lu, Vadim Dabravolski, Mark Dredze, Sebastian Gehrmann, Prabhanjan Kambadur, David Rosenberg, and Gideon Mann.
\newblock Bloomberggpt: A large language model for finance.
\newblock {\em arXiv preprint arXiv:2303.17564}, 2023.

\bibitem{cui2023chatlaw}
Jiaxi Cui, Zongjian Li, Yang Yan, Bohua Chen, and Li~Yuan.
\newblock Chatlaw: Open-source legal large language model with integrated external knowledge bases.
\newblock {\em arXiv preprint arXiv:2306.16092}, 2023.

\bibitem{dan2023educhat}
Yuhao Dan, Zhikai Lei, Yiyang Gu, Yong Li, Jianghao Yin, Jiaju Lin, Linhao Ye, Zhiyan Tie, Yougen Zhou, Yilei Wang, et~al.
\newblock Educhat: A large-scale language model-based chatbot system for intelligent education.
\newblock {\em arXiv preprint arXiv:2308.02773}, 2023.

\bibitem{zhang2023biomedgpt}
Kai Zhang, Jun Yu, Zhiling Yan, Yixin Liu, Eashan Adhikarla, Sunyang Fu, Xun Chen, Chen Chen, Yuyin Zhou, Xiang Li, et~al.
\newblock Biomedgpt: A unified and generalist biomedical generative pre-trained transformer for vision, language, and multimodal tasks.
\newblock {\em arXiv preprint arXiv:2305.17100}, 2023.

\bibitem{wang2023huatuo}
Haochun Wang, Chi Liu, Nuwa Xi, Zewen Qiang, Sendong Zhao, Bing Qin, and Ting Liu.
\newblock Huatuo: Tuning llama model with chinese medical knowledge.
\newblock {\em arXiv preprint arXiv:2304.06975}, 2023.

\bibitem{li2023huatuo}
Jianquan Li, Xidong Wang, Xiangbo Wu, Zhiyi Zhang, Xiaolong Xu, Jie Fu, Prayag Tiwari, Xiang Wan, and Benyou Wang.
\newblock Huatuo-26m, a large-scale chinese medical qa dataset.
\newblock {\em arXiv preprint arXiv:2305.01526}, 2023.

\bibitem{wang2019superglue}
Alex Wang, Yada Pruksachatkun, Nikita Nangia, Amanpreet Singh, Julian Michael, Felix Hill, Omer Levy, and Samuel Bowman.
\newblock Superglue: A stickier benchmark for general-purpose language understanding systems.
\newblock {\em Advances in neural information processing systems}, 32, 2019.

\bibitem{huang2023c}
Yuzhen Huang, Yuzhuo Bai, Zhihao Zhu, Junlei Zhang, Jinghan Zhang, Tangjun Su, Junteng Liu, Chuancheng Lv, Yikai Zhang, Jiayi Lei, et~al.
\newblock C-eval: A multi-level multi-discipline chinese evaluation suite for foundation models.
\newblock {\em arXiv preprint arXiv:2305.08322}, 2023.

\bibitem{hendrycks2020measuring}
Dan Hendrycks, Collin Burns, Steven Basart, Andy Zou, Mantas Mazeika, Dawn Song, and Jacob Steinhardt.
\newblock Measuring massive multitask language understanding.
\newblock {\em arXiv preprint arXiv:2009.03300}, 2020.

\bibitem{singhal2023large}
Karan Singhal, Shekoofeh Azizi, Tao Tu, S~Sara Mahdavi, Jason Wei, Hyung~Won Chung, Nathan Scales, Ajay Tanwani, Heather Cole-Lewis, Stephen Pfohl, et~al.
\newblock Large language models encode clinical knowledge.
\newblock {\em Nature}, 620(7972):172--180, 2023.

\bibitem{wang2023cmb}
Xidong Wang, Guiming~Hardy Chen, Dingjie Song, Zhiyi Zhang, Zhihong Chen, Qingying Xiao, Feng Jiang, Jianquan Li, Xiang Wan, Benyou Wang, et~al.
\newblock Cmb: A comprehensive medical benchmark in chinese.
\newblock {\em arXiv preprint arXiv:2308.08833}, 2023.

\bibitem{zhu2023promptcblue}
Wei Zhu, Xiaoling Wang, Huanran Zheng, Mosha Chen, and Buzhou Tang.
\newblock Promptcblue: A chinese prompt tuning benchmark for the medical domain.
\newblock {\em arXiv preprint arXiv:2310.14151}, 2023.

\bibitem{fei2023lawbench}
Zhiwei Fei, Xiaoyu Shen, Dawei Zhu, Fengzhe Zhou, Zhuo Han, Songyang Zhang, Kai Chen, Zongwen Shen, and Jidong Ge.
\newblock Lawbench: Benchmarking legal knowledge of large language models.
\newblock {\em arXiv preprint arXiv:2309.16289}, 2023.

\bibitem{dai2023laiw}
Yongfu Dai, Duanyu Feng, Jimin Huang, Haochen Jia, Qianqian Xie, Yifang Zhang, Weiguang Han, Wei Tian, and Hao Wang.
\newblock Laiw: A chinese legal large language models benchmark (a technical report).
\newblock {\em arXiv preprint arXiv:2310.05620}, 2023.

\bibitem{niklaus2023lextreme}
Joel Niklaus, Veton Matoshi, Pooja Rani, Andrea Galassi, Matthias St{\"u}rmer, and Ilias Chalkidis.
\newblock Lextreme: A multi-lingual and multi-task benchmark for the legal domain.
\newblock {\em arXiv preprint arXiv:2301.13126}, 2023.

\bibitem{islam2023financebench}
Pranab Islam, Anand Kannappan, Douwe Kiela, Rebecca Qian, Nino Scherrer, and Bertie Vidgen.
\newblock Financebench: A new benchmark for financial question answering.
\newblock {\em arXiv preprint arXiv:2311.11944}, 2023.

\bibitem{zhang2023fineval}
Liwen Zhang, Weige Cai, Zhaowei Liu, Zhi Yang, Wei Dai, Yujie Liao, Qianru Qin, Yifei Li, Xingyu Liu, Zhiqiang Liu, et~al.
\newblock Fineval: A chinese financial domain knowledge evaluation benchmark for large language models.
\newblock {\em arXiv preprint arXiv:2308.09975}, 2023.

\bibitem{lei2023cfbenchmark}
Yang Lei, Jiangtong Li, Ming Jiang, Junjie Hu, Dawei Cheng, Zhijun Ding, and Changjun Jiang.
\newblock Cfbenchmark: Chinese financial assistant benchmark for large language model.
\newblock {\em arXiv preprint arXiv:2311.05812}, 2023.

\bibitem{zhang2019adversarial}
Jiliang Zhang and Chen Li.
\newblock Adversarial examples: Opportunities and challenges.
\newblock {\em IEEE transactions on neural networks and learning systems}, 31(7):2578--2593, 2019.

\bibitem{yang2023baichuan}
Aiyuan Yang, Bin Xiao, Bingning Wang, Borong Zhang, Ce~Bian, Chao Yin, Chenxu Lv, Da~Pan, Dian Wang, Dong Yan, et~al.
\newblock Baichuan 2: Open large-scale language models.
\newblock {\em arXiv preprint arXiv:2309.10305}, 2023.

\bibitem{bai2023qwen}
Jinze Bai, Shuai Bai, Yunfei Chu, Zeyu Cui, Kai Dang, Xiaodong Deng, Yang Fan, Wenbin Ge, Yu~Han, Fei Huang, et~al.
\newblock Qwen technical report.
\newblock {\em arXiv preprint arXiv:2309.16609}, 2023.

\bibitem{sun2021ernie}
Yu~Sun, Shuohuan Wang, Shikun Feng, Siyu Ding, Chao Pang, Junyuan Shang, Jiaxiang Liu, Xuyi Chen, Yanbin Zhao, Yuxiang Lu, et~al.
\newblock Ernie 3.0: Large-scale knowledge enhanced pre-training for language understanding and generation.
\newblock {\em arXiv preprint arXiv:2107.02137}, 2021.

\bibitem{zhang2019bertscore}
Tianyi Zhang, Varsha Kishore, Felix Wu, Kilian~Q Weinberger, and Yoav Artzi.
\newblock Bertscore: Evaluating text generation with bert.
\newblock {\em arXiv preprint arXiv:1904.09675}, 2019.

\end{thebibliography}

\end{document}